\documentclass[conference]{IEEEtran}

\usepackage{epsfig}
\usepackage{subcaption}
\usepackage{calc}
\usepackage{amssymb}
\usepackage{amstext}
\usepackage{amsmath}
\usepackage{amsthm}
\usepackage{multicol}
\usepackage{pslatex}
\usepackage{apalike}
\usepackage[bottom]{footmisc}

\begin{document}

\title{Automatic Defect Detection in Sewer Network using Deep Learning based Object Detector}

\author{
    {\Large Bach Ha$^1$, Birgit Schalter$^2$, Laura White$^1$, Joachim Köhler$^1$} \\
    {\small $^1$NetMedia, Fraunhofer IAIS, Sankt Augustin, Germany} \\
    {\small $^2$Dr.-Ing. Pecher und Partner Ingenieurgesellschaft mbH, Berlin, Germany} \\
    {\small \texttt{\{duc.bach.ha, laura.white, joachim.koehler\}@iais.fraunhofer.de, birgit.schalter@pecherundpartner.de}}

}
\maketitle

\begin{abstract}
  Maintaining sewer systems in large cities is important, but also time and effort consuming, because visual inspections are currently done manually. 
  To reduce the amount of aforementioned manual work, defects within sewer pipes should be located and classified automatically.
  In the past, multiple works have attempted solving this problem using classical image processing, machine learning, or a combination of those.
  However, each provided solution only focus on detecting a limited set of defect/structure types, such as fissure, root, and/or connection.
  Furthermore, due to the use of hand-crafted features and small training datasets, generalization is also problematic.
  In order to overcome these deficits, a sizable dataset with 14.7 km of various sewer pipes were annotated by sewer maintenance experts in the scope of this work.
  On top of that, an object detector (EfficientDet-D0) was trained for automatic defect detection.
  From the result of several expermients, peculiar natures of defects in the context of object detection, which greatly effect annotation and training process, are found and discussed.
  At the end, the final detector was able to detect $83\%$ of defects in the test set; 
  out of the missing $17\%$, only $0.77\%$ are very severe defects.
  This work provides an example of applying deep learning-based object detection into an important but quiet engineering field.
  It also gives some practical pointers on how to annotate peculiar "object", such as defects.
\end{abstract}


\section{\uppercase{Introduction}}
\label{sec:introduction}

Sewer systems in large cities require continuous enormous amounts of maintenance; for example, in Berlin ~650 km of sewer pipes must be inspected each year. 
This inspection process is done by domain experts viewing video recordings of pipe interiors and marking defects manually.
Consequently, such a process is time-consuming, tedious, and error-prone.
In order to reduce the amount of necessitated manual effort, multiple previous works attempted to automatically defects using classical image processing methods, and rudimentary machine learning on hand-crafted features \cite{Makar1999DiagnosticTF}.
As a result, each work is limited to only one or two specific type of defects.
Generalization to changes in appearance of defects and background is also another potential issue.
While classical image processing based works' result are promising, they are not yet enough for practical use on the field.
On the other hand, modern deep-learning based works have also been developed as an attempt to solve this task.
While delivering good results, networks from these works are trained and evaluated on relatively small datasets \cite{cheng2018automated,kumar2020deep}.
Consequently, generalization, or how networks behave to variation in inputs, is a potential problem.
In order to deal with the aforementioned deficits, this work aimed to develop an automatic deep learning (DL) based defect detector, which is trained and evaluated on a new sizable and varied dataset.
This detector's architecture is Efficient-Det D0 \cite{Tan2020EffDet}.
Beside automatic evaluation, the final evaluation is also done by expert engineers in the field, thus providing a thorough and practical report on the network's performance. 
This paper is divided into five main sections.
Related work provides a brief overview on the current status of automatic defects detection in sewer systems, and deep learning based vision object detection.
The second section explains methodology, as well as accompanying problems for data acquisition, annotation, network training, and evaluation.
Next are the result section, future works section, and finally the conclusion.

\section{\uppercase{Related Work}}
\label{sec:related}
This section gives more details on previous works in the field of sewer maintenance that attempted to solve the same problem and their deficits.
In addition, a short overview on deep-learning based object detections is also provided.

\subsection{Sewer Maintenance}
\label{subsec:sewer}
In order to efficiently inspect inaccessible sewer systems multiple non-destructive diagnostic methods have been developed since at least 1981 \cite{Makar1999DiagnosticTF}.
While systems making use of sensors, such as ground penetrating radars, ultrasound, laser-scanner \cite{Duran2002,bailey2011real} have been developed and applied, closed-circuit television (CCTV) based methods \cite{Makar1999DiagnosticTF} are more popular due to the intuitive nature of the data, which can be manually analyzed by technicians.
Earlier CCTV based methods \cite{Moselhi1999,Sinha2000AutomatedUP,Sinha2002,Yang2008automated} made use of classical image processing such as edge detection, or boundary segmentation to detect the present of defects on pipes' inner surface, and perform feature extraction.
After that, classification of defects are done either by using heuristic from fixed extracted features (size, diameters, etc.) \cite{Sinha2000AutomatedUP,Muller2009,huynh2015anomaly,tung2015segmentation} or various machine learning technics, for example neural network (NN) \cite{Moselhi1999,Hassan2019102849}, fuzzy-neural network \cite{Sinha2002}, radial basis network (RBN), and support vector machine (SVM) \cite{Yang2008automated,hengmeechai2013automated}.
While producing results, these classical/hybrid methods require careful choices and configurations of image preprocessing methods based on defined target types of defects.
In consequence, each resulting solution is limited to one or two very specific types of defect (i.e. fissure, joint open, root).
Furthermore, hand-crafted configurations make the solution inflexible to variations in environments, and defects.
Another problem is that most of them depend on edge detection, which make defects with similar forms, for example fissure and root, not differentiable.
In recent years, such shortcomings could potentially be solved using artificial neural networks (ANN).
Although neural networks were already mentioned and used, these networks are still working on top of hand-crafted feature vectors, thus continue to be limited by the original configuration.
An explanation for such decisions was the lack of computational power \cite{Moselhi1999,Yang2008automated} at the time.
A more recent paper \cite{Hassan2019102849} has been able to perform classification directly on RGB input images, however a technician is required to control the camera view and locate potential defects.
Recently, thanks to advancements in hardware, concerns over performance issues of neural networks are less relevant.
Which leads to the existence of multiple methods using complex deep neural networks (DNN) \cite{kunzel2018automatic,cheng2018automated,kumar2020deep,wang2021towards}, which not only classify but also localize defects simultaneously.
These newer methods feed camera images directly into a FasterRCNN \cite{cheng2018automated,kumar2020deep,wang2021towards}, or YOLOv3 \cite{kumar2020deep} object detection network, which produces bounding boxes representing location and classification of defects.
However, these deep-learning based methods used relatively small datasets, which only focus on a small set of defects (fissure, root, infiltration, deposit), for training and evaluation.
Another problem is that, these earlier network only provided a perspective view of part of detected defects or complex defect system, which would not allow easy automatic measurement and damage assessment. 
Another type of DNN that performs image segmentation, namely FRRN \cite{kunzel2018automatic}, was also used in an effort to detect sewer pipes' defects.
This method however projects raw video images into a single 2D unrolling of the pipe, similar to \cite{Muller2009}, which is then fed to the FRRN for segmentation.
Regardless, the segmentation network from this work has problem with overlapping defects.

\subsection{DL based Vision Object Detection}
Detecting objects in RGB images produced by CCTV is a suitable task for deep learning based object detectors.
These detectors are a subset of DNN, which specialize in localizing and classifying objects on visual data, such as RGB images.
There are currently three main groups of such detectors, namely two-stage detectors, one-stage detectors \cite{Zhao2019}, and recently developed transformer-based detectors.

Two-stage detectors are networks such as FastRCNN \cite{Girshick15FastRcnn}, or FasterRCNN \cite{Ren2015FasterRcnn}.
These networks function somewhat similar to classical image processing methods, by first locating region of interests (ROI), each of which is the location of a potential object.
Next, classification is performed on each ROI to get the object class. 
Thanks to the separation of localization and classification functionality, two-stage detectors can reliably provide accurate object location and class.
Two-stages networks' training process are also more stable and easier to control, because the localization part and classification part can be trained separately.
In return, they are slower and require a more complicated training process.

On the other hand, one-stage detectors, such as networks from the YOLO-family \cite{Redmon2016Yolo,Redmon2018Yolo}, Single-Shot-Detector (SSD) \cite{Liu2016SSD}, or the EfficientDet-family \cite{Tan2020EffDet}, trade training stability for better performance and a simpler training process by performing both localization and classification simultaneously.
Being a unified calculation graph allowed these networks to be optimized on lower software and hardware level, thus greatly increasing their processing speed.
The same characteristic however makes it difficult to improve the localization and classification separately, since both parts have to be trained together.
Earlier one-stage detectors generally have lower accuracy than two-stage detectors, however this is no longer always true for newer iterations \cite{Zhao2019,Wang2022Yolo7}.

The third and final group, transformer detectors are recently introduced networks, which makes use of attention mechanism instead of pure convolutional layers. 
One of the first and most famous network of this group is DETR \cite{Carion2020DETR}.
Since then transformer-based detectors have consistently produced highly accurate detections, and one of them \cite{wang2022internimage} is currently at the top of the COCO benchmark \cite{Lin2014COCO}.
However, these networks are extremely large and much slower in comparison to networks in the other two groups.
Furthermore, the required amount of data for training them is also very high.

\section{\uppercase{Method}}
\label{sec:method}

\begin{figure}[!h]
  \centering
   \includegraphics[width=\linewidth]{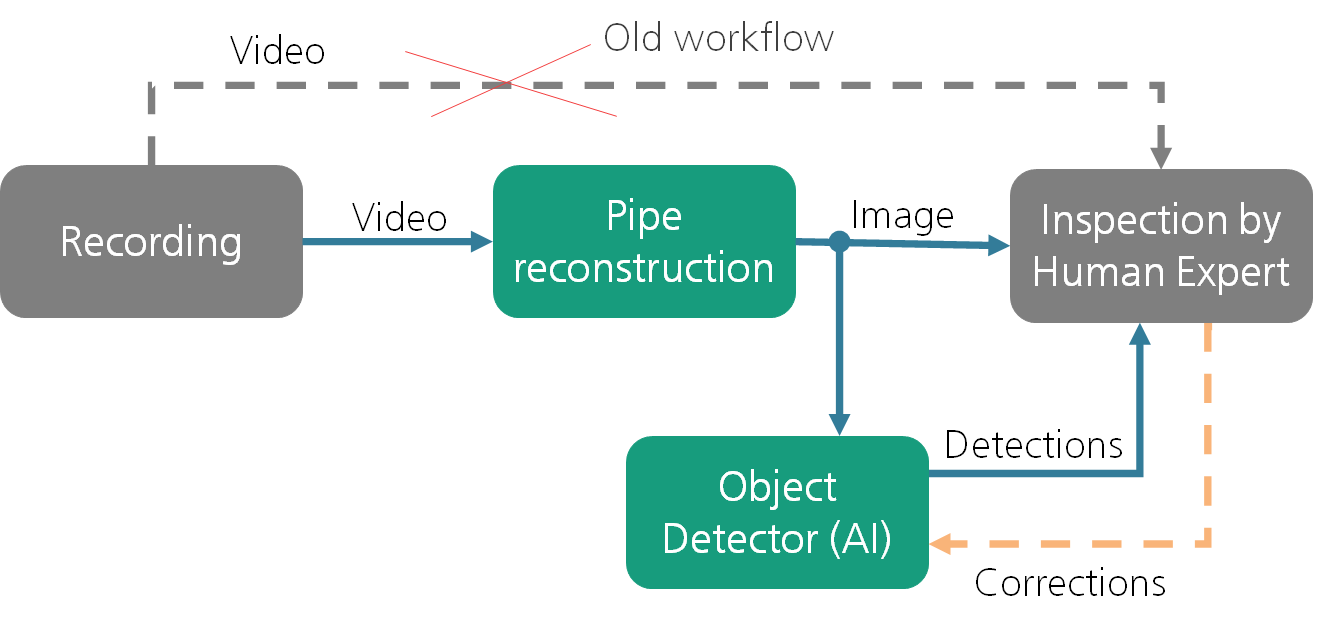}
  \caption{New inspection workflow with automatic object detector integration.}
  \label{fig:workflow}
\end{figure}

\begin{figure*}[!h]
  \centering
   \includegraphics[width=\linewidth]{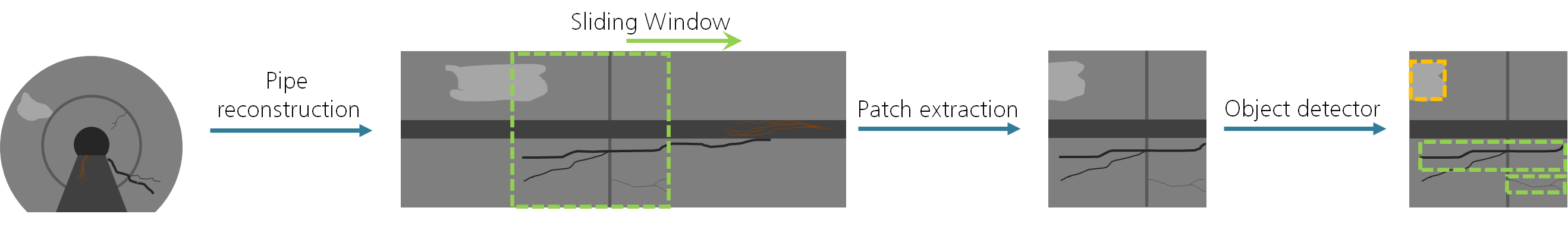}
  \caption{Visualization of data within the pipeline.}
  \label{fig:dataflow}
\end{figure*}

To satisfy the use-case, the to be designed solution has to adhere to the following requirements.
Firstly, the system must allow technicians to pinpoint the location of defects, as well as providing them with an overall and clear view of defects.
Secondly, produced detections should allow an accurate measurement of defect size and length (i.e. of fissures, and root).
The third requirement is that the annotation process for training data should not be complicated and extremely time-consuming.
Next, the system should require minimal controlling effort and time on site, avoiding closing off the street for a long period of time.  
Finally, processing of recorded video data should be efficient and done in a reasonable amount of time without the need of expensive computation clusters.

With the aforementioned requirements in mind, the following design choices were made.
In order to fulfill the first and second requirements, the network will not work directly on video frames, but on unrolled 2D projections of pipes' inner surface, which was similarly done in \cite{Muller2009,kunzel2018automatic}.
This method also reduces the amount of images, that have to be processed by the neural network, because near identical frames from camera recordings are eliminated.
The third and fourth requirements rule out the application of segmentation neural networks in \cite{kunzel2018automatic}.
Because of the complexity and required accuracy of segmentation masks, labeling at pixel level is very time-consuming \cite{CordtsORREBFRS16} in comparison to drawing bounding boxes over defects.
Segmentation networks are also generally slower and require more computational resources in comparison to bounding box (BB) based detectors.
Therefore, BB-based object detectors would be the more suitable choice.
The last two requirements directly exclude transformer based networks, and favor single-stage detectors over two-stages detectors;
with the type of detector narrowed down, there is still the choice of the specific network architecture, such as SSD, RetinaNet \cite{Tsung2017FocalLoss}, YOLO-Family, or EfficientDet.
EfficientDet was chosen, because at the time it was the current state-of-the-art \cite{Tan2020EffDet}.
However, it should be noted that newer versions of YOLO (YOLOv5, YOLOv7 \cite{Wang2022Yolo7}) are also promising.

\subsection{Dataset Acquisition and Annotation Problems}
\label{subsec:data}
For the purpose of training the chosen neural network, a sewer defects detection dataset is constructed from maintenance fisheye videos of Berlin's sewer system.
Before anything else is done, each video is unrolled and stitched into a single $W\times1200$ RGB image using the method described in \cite{kunzel2018automatic}. 
The width $W$ of each output image depends on the length of each inspected pipe.
With the RGB input for the network secured, the next important step is obtaining bounding box annotations.
For this step, a set of 9 common defects and 1 structure element similar to \cite{kunzel2018automatic} is defined. 
Defects include settled deposits (BBC), break/collapse (BAC), deformation (BAA), obstacle (BBE), angular displaced joint (BAJ C), surface damage (BAF), horizontal displaced joint (BAJ B), fissure (BAB), and root (BBA).
While structure element only consist of one class: connection (BCA).
Name of all defects/structure are taken from the English translation of the German standard: "DIN EN 13508-2:2011-08" \cite{DIN2011}.
While the accompanied letter codes are from the Euronorm.
In total, the dataset includes 14.7 km of annotated sewer pipes.
\begin{figure}[!h]
  \centering
   \includegraphics[width=\linewidth]{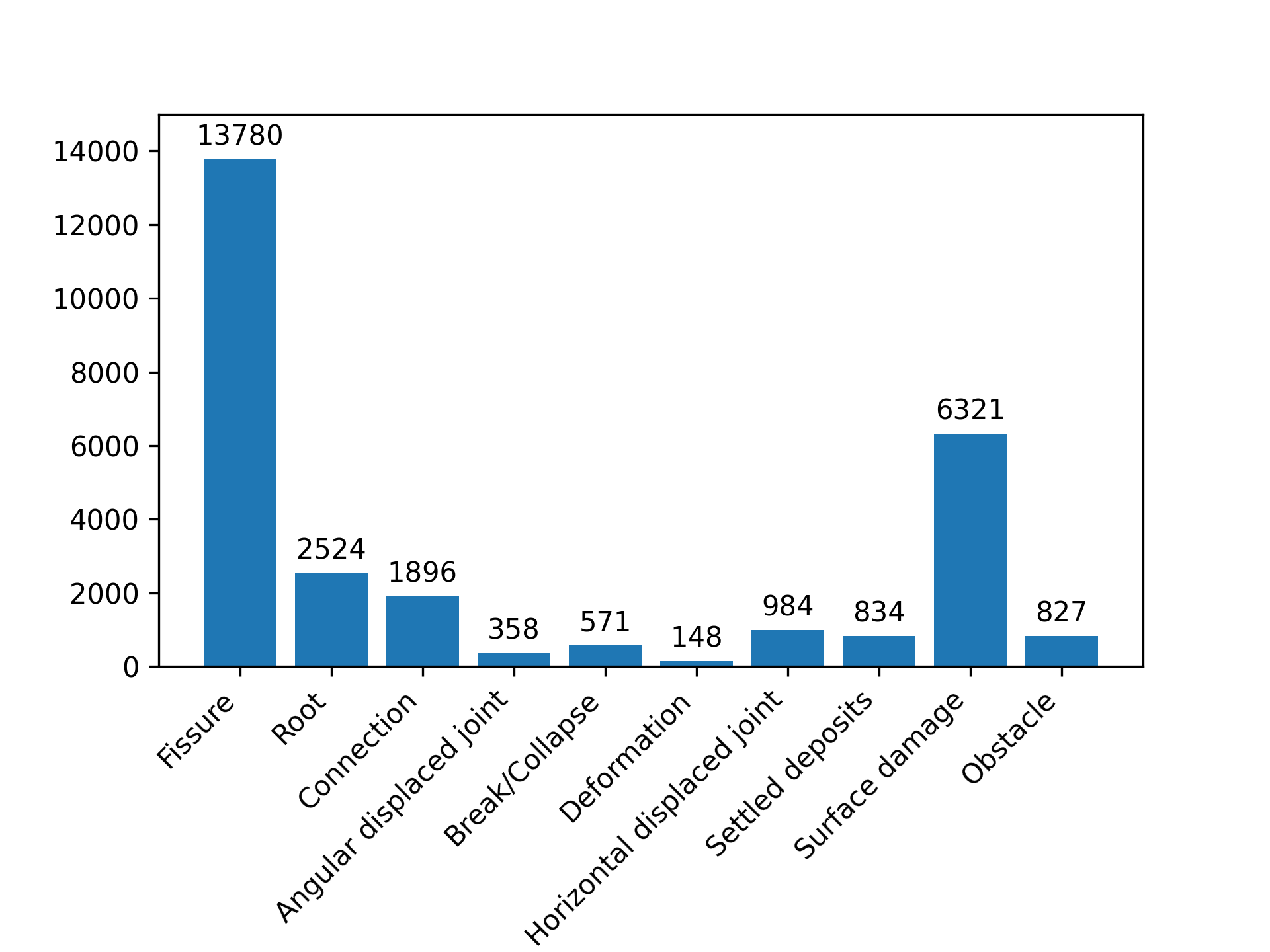}
  \caption{Distribution of defect types.}
  \label{fig:class_dist}
\end{figure}

\begin{figure}[!h]
  \centering
   \includegraphics[width=\linewidth]{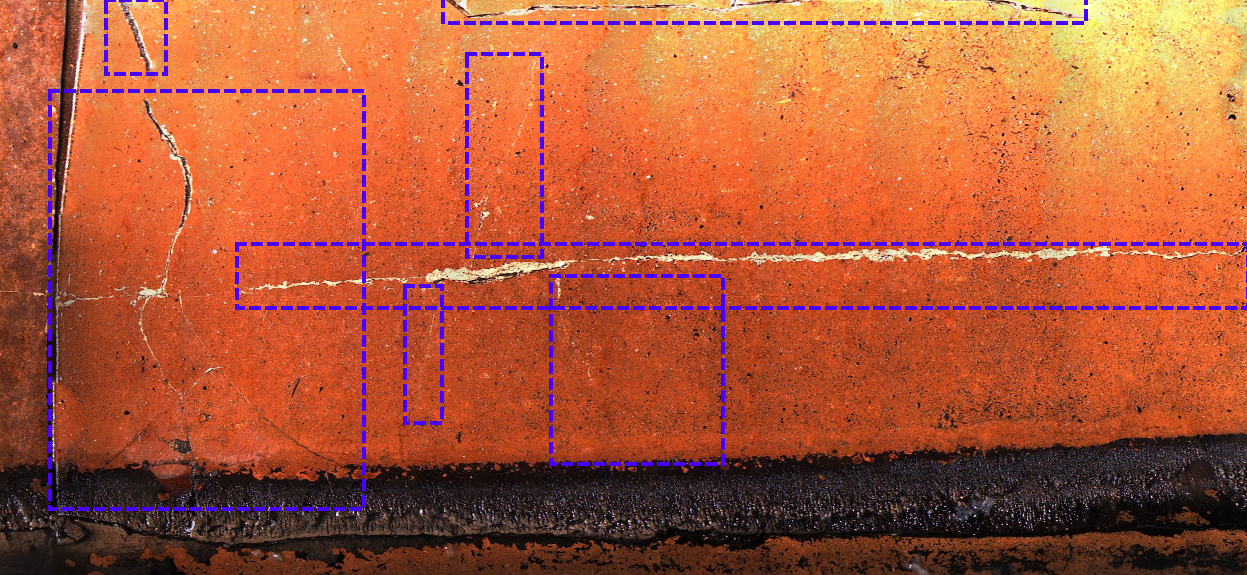}
   \includegraphics[width=\linewidth]{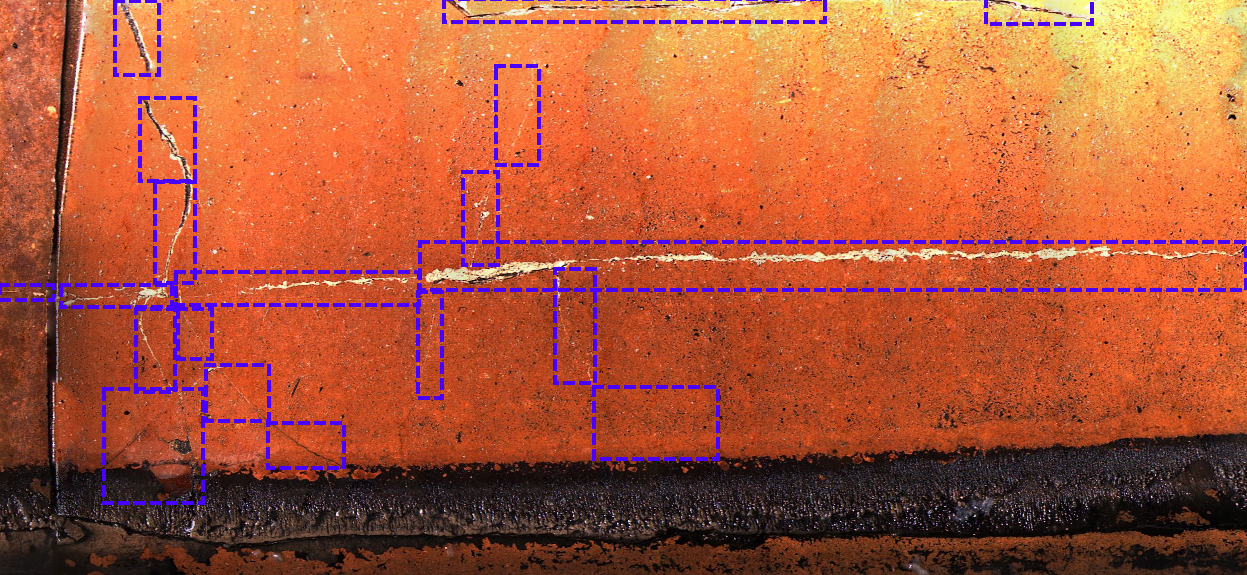}
  \caption{Examples for how fissures could be labeled, either coarsely (top) or finely (bottom).}
  \label{fig:2vis}
\end{figure}

Beside the known large amount of effort needed to annotate a machine learning dataset, the defects in this dataset also introduce their own additional problems.
Although defined as an object detection dataset, unlike COCO\cite{Lin2014COCO} or KITTI\cite{Geiger2012Kitti} datasets, the task of drawing a box over an object and assigning a class to it for this dataset is not as straight forward, which cost a lot of time and necessitated multiple iterations of annotations.
The main cause is the uncommon and sometime ambiguous nature of defects within sewer pipes. 
Unlike natural objects such as cat, dog, or car, which mostly require only common sense, defects require training and experience to be accurately diagnosed.
This means that the labeling process has to be carried out only by maintenance experts to ensure higher annotations quality.
With that in mind, in order to accelerate the labeling process, the first iteration of annotation was done by letting experts work in parallel on different sewer pipes.
That was a wrong decision that lead to the failure of the first iteration of annotation, in which produced labels for defects are so inconsistent that neural network become more confused after the training.
The first problem is that, even to experts, some defects are still ambiguous.
In another word some experts might classify an "object" as a defect, while others do not, which further confirms the finding made in \cite{Mueller2006}.
The second problem is deciding on a way to draw bounding boxes consistently, which sensibly represents relevant "object". 
Specifically, this is a problem with annotating fissure, root, and surface damage.
Thinking back to objects in COCO or KITTI; although color, size, and orientation of these objects are varied, they all have certain fixed shape, form, or ratio and can be separated into single instances (for example, a cat, a dog, a bike, ...).
On the other hand, fissure, root, and erosion exist mostly in form of clusters, which can be of any shape, size, and density.
Therefore, it is not trivial to intuitively define an "instance" for these defects, which can then be surrounded by a box. 
Without a consistent and unified guideline, each expert labeled these clusters with different levels of coarseness, further exacerbating the level of inconsistency in the dataset. 
Therefore, it is clear that common labeling rules must be set.
From analyzing existing labels, there are two possible directions to annotate such defects.
The first direction is to coarsely label each entire region of connected defects within a single bounding box. 
The other direction is to finely divide each cluster into multiple smaller "instances", each of which is labeled by a bounding box.
Here, smaller instance is loosely defined as the largest continuous segments of defects, which can be fitted into a tight bounding box, that has high "defect to background" area ratio. 
Examples in figure \ref{fig:2vis} help show the difference between these two aforementioned labeling directions.

Although finely divided labels do seem excessive in the example, this inconvenience is outweighed by more precise detections.
Preliminary training and testing on a subset of the dataset shows that network trained with coarse labels is unable to detect smaller and finer fissures, roots, and erosion areas.
In contrast, networks trained with finely divided clusters are capable of detecting not only smaller defects but also larger defect clusters.
While detection of such large clusters are represented with a lot of smaller overlapping bounding boxes, that is still much better than completely missing defects, especially when these smaller boxes can later be combined using various post-processing methods.

With all this hindsight, in order to deal with the ambiguous factor of defects and variations in labeling style, a second round of annotations was done following a stricter labeling guideline, with all experts working together as a single group and vote on every annotated defect.
As a result, the annotations from the second round are empirically more consistent.
Finally, with large problems of the dataset sorted out, the next step is to train the neural network.

\subsection{Neural Network Training}
This subsection describes main steps and corresponding configuration to train the required detector.
As reasoned in the beginning of this section \ref{sec:method}, a network from the EfficientDet family \cite{Tan2020EffDet} was trained to detect defects and structures within sewer pipes, specifically EfficientDet-D0 was used.
The training follows a standard procedure of multiple steps.

Firstly, the whole dataset is split into a train set and a test set.
For large public datasets, splits are often already chosen and provided \cite{Lin2014COCO,Geiger2012Kitti}.
When that is not the case, splits are generally generated randomly from the whole dataset. 
For such a new and self labeled dataset, random split was a decent choice, which costs little time and effort.
However, results of the first few cross validations on random test sets shown large and chaotic variations, where the network achieves really high accuracy in some rounds while completely failing to recognize severe defects in others.
After further inspecting those automatic splits, it was found that due to difference in rarity of each defect type and sewer pipe conditions, characteristics of each random split changed drastically.
For example, there are splits, where the majority of break/collapse or obstacle are concentrated in the test set, thus the network is tested on defects that it was hardly trained on.
On the other hand, the test set could mostly consist of healthy pipes, in which the network easily detected most of the scattered defects.
In both cases, these networks are all incomparable, and the evaluation untrustworthy due to a lack of stable baselines.
Therefore, a set of ten representative sewer pipes were instead manually chosen and set aside for evaluation.
For clarity, representative means that these pipes must contain all the relevant defects and structure with good variation.
Furthermore, pipes with defect free sections are also included to make sure that the resulting network does not wrongly mark healthy pipe sections as damaged.

With data splits sorted out, the second step is preprocessing, some light data augmentation, and feeding the neural network with RGB images.
This however cannot simply be done like with COCO or KITTI because of the images' length.
Unrolled RGB images of pipes have resolution of around 20000x1200 pixel to 150000x1200 pixel, which are too large and simply result in Out-Of-Memory (OOM) error when fed directly into a detection network.
Therefore, the solution is a simple sliding window approach with each patch having the size of 1200x1200 pixel.
There is also a 50\% overlap between each window, so that each defect is likely to be in full view of the network.
After the original image is cut into multiple patches, data augmentation is then applied to each patch separately.
Each input patch and their corresponding annotation have a 25\% chance to be flipped up-down and another separate 25\% chance for left-right flipping.
This minimal augmentation helps increase the variety of the dataset without introducing too much artificial artifacts, which might have unknown adverse effects on new unknown data.
One final step before the network gets to see the data is down-sampling from 1200x1200 to 640x640.
This was done so that the detector can be trained with a batch size larger than 1.
Another added benefit is that the network also requires less time for inference and training.
While accuracy could suffer due to lowered resolution, a preliminary test shown negligible change after down-sampling. 

The third step is configuring the training with plausible values of hyperparameters, that are suitable for the dataset and available hardware.
These hyperparameters can be roughly divided into two groups: network hyperparams and trainer hyperparams.
However, it should be noted, that both groups are not independent but greatly affect each other.
To the first group, the network hyperparams are, for example, weight initialization options, or batch normalization \cite{IoffeS2015bn} configurations.
Within this group, the most important hyperparam is weight initialization, which was set to use weights pre-trained on COCO \cite{Tan2020EffDet}.
While simpler methods, such as variations of random initialization exist, they are better suited for new unknown architectures \cite{HeZR015Kaiming}.
Since the chosen EfficientDet-D0 is already established, and it is already shown that transfer learning greatly improves the final trained network \cite{Tan2019EffNet}, pre-trained weights initialization is more logical.
To be more specific, pre-trained weights allow networks to reuse proven learned feature extractors.
This spares the training from the earlier divergence-prone phase, which also requires a large amount of data.
As a result, such initialization is especially useful for small datasets, that do not necessarily have enough samples to properly stabilize its feature extractors from scratch.
This is still the true for the defect dataset in this work, despite the fact that this dataset's domain is entirely different from that of normal datasets, namely COCO or KITTI (normal scenes vs. sewer pipe's inner surfaces).
With the network configured, trainer's hyperparams are next.
For this group, training hardware, especially the graphical processing unit (GPU), has a lot of says in the configuration.
In this case, an Nvidia GTX Titan X GPU with 12 GB of random access memory (RAM) was used.
Thanks to the previous down-sampling step, a batch size of 3 is set for the training.
The optimizer is Adam \cite{kingma2014adam} with default parameters as suggested in its original paper.
While there other optimizers like stochastic gradient descent (SGD) \cite{Ruder2016Optimizer} and its variants exist, Adam is shown to keep the training process stable with little to no additional hyperparameter configuration \cite{kingma2014adam,Ruder2016Optimizer}, which is especially important when working with new unknown data.

The fourth and longest step in training a neural network is waiting.
The training process is supervised using a combination of loss log, mean average precision (mAP) \cite{Lin2014COCO} calculation, ROC-curve, sample outputs between epochs, and a custom practical metric (see section \ref{subsec:eval}).
This enables early stopping of training, either manually or automatically if performance reaches a plateau or worsens.
On the aforementioned GPU, with a total of around 71300 training patches, each training took approximately seven days before being stopped. 

The fifth and final step is evaluating the trained network on the pre-defined test set from section \ref{subsec:data}.
Evaluation provides a close estimation of trained networks' quality, which enables informed decision on whether the training is finished, or what has to be done in the next training cycle to deal with network's deficit. 
In object detection, the standard procedure for this step is running newly trained networks on test set to produce predictions.
After that, the quality of these predictions, and of the networks, are then determined automatically using a quantified metric such as mAP.
While higher mAP generally means better network, there are cases where the corresponding predictions are not ideal despite high mAP \cite{Redmon2018Yolo}.
Other metrics such as precision, recall, and the harmonic f1 score also work well as indicators for network's quality.
In addition to automatic mAP calculation, manually examining predictions of a random subset of the test set is often done to ensure the quality of trained networks.
However, this standard procedure is not directly applicable into this work, forcing some changes, which make it more suitable for practical applications. 
The deficit and corresponding solutions for the evaluation procedure would be discussed in the next subsection \ref{subsec:eval}.

\subsection{Problem with mAP and Network evaluation}
\label{subsec:eval}
\begin{figure}[!h]
  \centering
   \includegraphics[width=\linewidth]{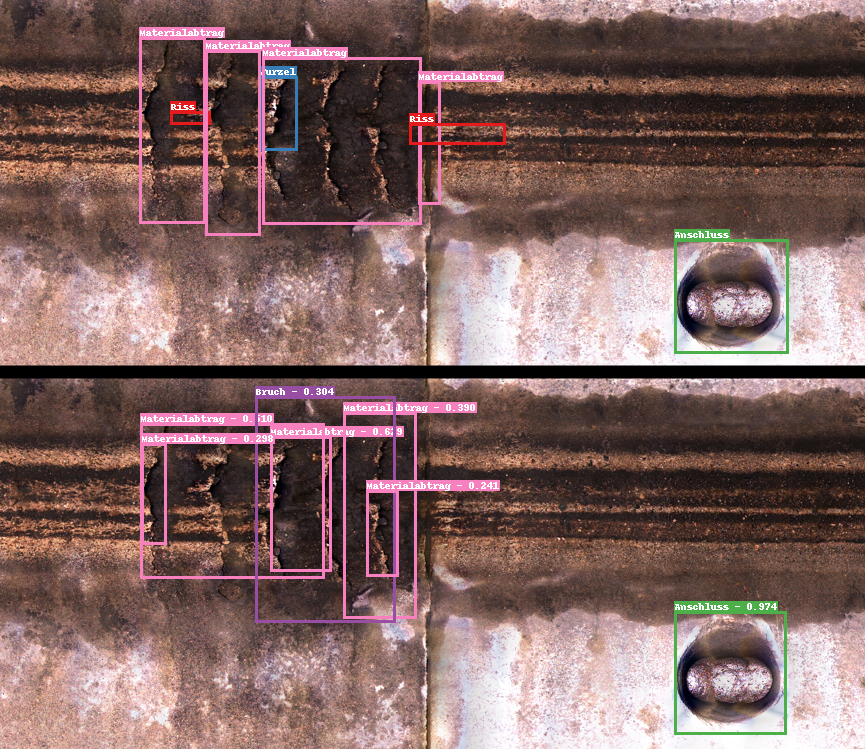}
  \caption{Differences in experts' annotations (top), and the networks' detections (bottom) of formless defects, in this case surface damages, which are marked with pink boxes}
  \label{fig:labelvsdet}
\end{figure}
This section provides details on why the standard evaluation is not suitable for this use case, and how to it was dealt with.
The first sight of problem was detected by simply following the established procedure, calculated mAP for all best trainings are around only 5 mAP.
Such low score normally indicates one or more of the following problems, namely bad network architecture, misconfiguration of input images and labels, and/or peculiar problems of the dataset.
Bad network architecture is unlikely the reason, since EfficientDet is able to handle large and complex dataset \cite{Lin2014COCO}.
Careful review of the preprocessing pipeline confirms no misconfiguration of input data.
With other potential sources of problem ruled out, the remaining and most likely the reason for low mAP score is the dataset itself.
Given all taken precautions during data gathering, as mentioned in section \ref{subsec:data}, this raises the question of what exactly the problem with the dataset would be.
A step to answering this question is manually checking the network's detections and comparing those with the annotation.
This reveals a downside of using bounding boxes (BB), and by extension BB-based mAP calculation, for denoting formless and cluster-like defects (.i.e fissure, root, surface damage).
Manual check confirms that the network is indeed able to detect those defects and correctly denote defects areas with multiple boxes, however these generated bounding boxes do not have the exact configuration of the ground truth boxes on the same damaged areas.
For example, network draws a single large box while the annotation has multiple smaller side by side boxes, or vice versa (see figure \ref{fig:labelvsdet}).
This is especially bad for long fine fissures or roots that run diagonally across the surface. 
Therefore, from a practical point of view, mAP does not correctly represent the network's quality, when such formless defects are concerned.
As a side note, while mask annotation could have solved this problem, it does have other undesirable downsides, as mentioned in the start of section \ref{sec:method}.
Another problem with mAP is that, it lacks an intuitive baseline to compare against.
The current highest achievable mAP on the COCO dataset is 65.4 mAP \cite{wang2022internimage}, which only shows that, that new network is better than EfficientDet-D0 (at 34.6 mAP) at producing predictions closer to the ground truths.
34.6 mAP does not mean that EfficientDet-D0 can only correctly detect 34.6 out of every 100 objects; that number is evidently higher \cite{Tan2020EffDet}.
The SoTA 65.4 mAP also does not necessarily mean a double in accuracy in comparison to EfficientDet-D0.
Unlike, for example, percentage where 50\% means the network handles half of given tasks correctly.
All in all, mAP has limited use in practical applications.
Now that the problem has been identified, the next part will go into the solution.

With the SoTA metric deemed less suitable for the job, it was decided that the final round of evaluation must be done manually by maintenance experts.
While time and effort costly, this eliminates any uncertainty regarding a network's quality.
Ironically, the smaller dataset size helps make this task more manageable.
However, it is not realistic to expect the experts to examine every single experimental training with different hyperparameter configurations, because that would be a lot of unnecessary work and the waiting time for results would be too long. 
Therefore, an automatic metric for internal quality estimation during experimentation, which mAP was supposed to be, need to be created.
In practice, the exact location, and amount of defects on pipes' inner surface is not required by the experts.
For them, it is enough, and more practical to know on which meters of pipe defects exist, and of what type.
Based on suggestions from the experts, and publication \cite{Berger2020Zustand}, it was determined that running kilometers/ meters is a standard baseline for metrics in the field of sewer sanitation.
Hence, for practical purposes, the internal metric would be using running meters as base.
To calculate this metric, each pipe is first divided into multiple $600x1200$ chunks. 
Each chunk would then be evaluated separately and would be marked as true positive (TP), false positive (FP), true negative (TN), or false negative (FN) depending on bounding boxes predictions within the chunk.
A chunk is deemed as TP if it contains at least one bounding box with the correct class of one of the defects in that chunk; 
the exact location and total number of predictions would not be taken into consideration.
A FP is given when the network put bounding boxes in a defect-free chunk.
A chunk is marked as TN when the network does not produce any bounding boxes in a defect-free chunk.
A TN is asserted when the network failed to produce any bounding boxes in a chunk with actual defect;
this is also the worst case that must be minimized.
After all chunks are evaluated, standard statistics, such as accuracy, precision, or recall can be easily calculated.
\begin{figure}[!h]
  \centering
   \includegraphics[width=\linewidth]{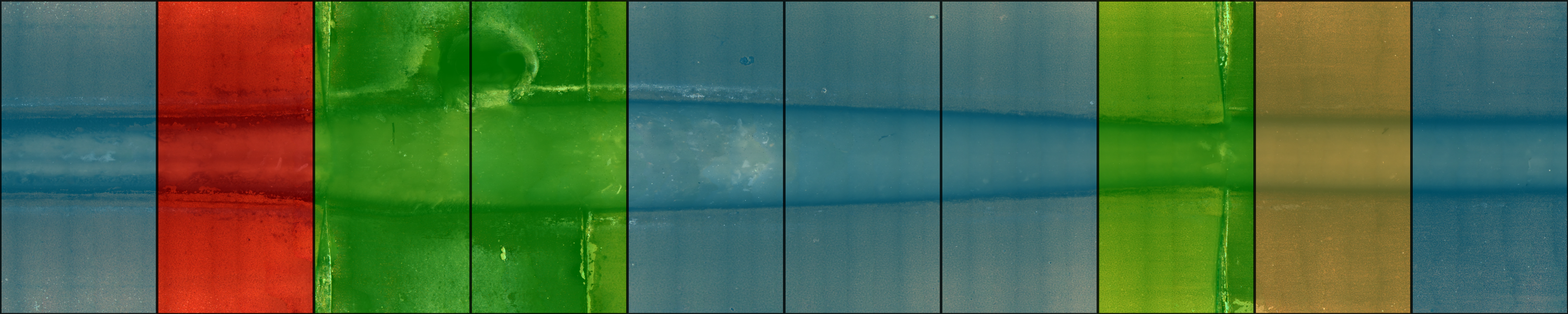}
  \caption{An example of the running meter metric; TP: Green, TN: Blue, FP: Yellow, and FN: Red.}
  \label{fig:running_metric_ex}
 \end{figure}

While this metric is clearly too lax in comparison to mAP for SoTA object detection, it is for experts in field of sewer system more understandable and useful.
Intuitively, what this metric says is, that within all pipe sections of $N$ meters, $X\%$ of them have defects, which are most likely of the following $Y, Z$ type.
With that knowledge, experts could then directly go to the marked section to perform thorough inspection, thus making the final decision on whether that pipe section is to be fixed or not.

To sum up, in this work, each trained network would first be automatically evaluated using the running meters metric.
The promising ones are then sent to experts for the final and real evaluation.
With the method and metrics for evaluation defined, the next section would describe the final defect detection result.

\section{\uppercase{Result}}
\begin{figure*}[!h]
  \centering
   \includegraphics[width=\linewidth]{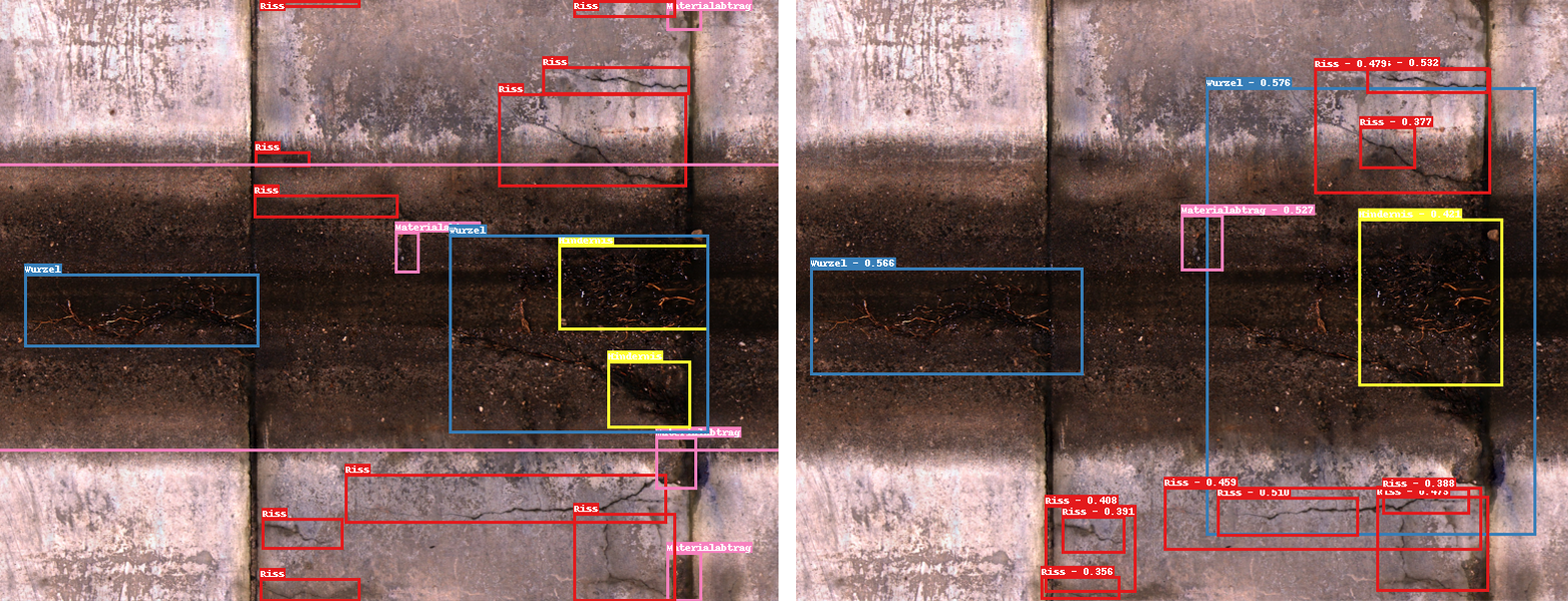}
  \caption{Visualization of experts' annotations (left) and the networks' detection (right) on a pipe section with multiple defects.}
  \label{fig:labelvsdet_final}
\end{figure*}

\begin{table}
  \centering
  \caption{Expert's evaluation on test set; precision and recall for each defect/structural class. Statistics calculated from evaluation results provided by the Dr.-Ing. Pecher und Partner Ingenieurgesellschaft mbH}
  \begin{tabular}{|l|l|l|r|}
    \hline
    Defects/Struct.  & Precision & Recall & N object \\ \hline
    Fissure          & 0.6697    & 0.6854 & 154      \\ \hline
    Root             & 0.7182    & 0.8778 & 129      \\ \hline
    Connection       & 0.9565    & 1.     & 51       \\ \hline
    Ang. dis. joint  & 1.        & 0.04   & 25       \\ \hline
    Break/collapse   & 0.037     & 0.5    & 2        \\ \hline
    Deformation      & 0.        & 0.     & 3        \\ \hline
    Hor. dis. joint  & 0.5659    & 0.9626 & 4        \\ \hline
    Settled deposit  & 0.        & 0.     & 43       \\ \hline
    Surface damage   & 0.8903    & 0.8406 & 80       \\ \hline
    Obstacle         & 0.6875    & 0.7952 & 78       \\ \hline
    All average      & 0.5525    & 0.5702 & 569      \\ \hline
  \end{tabular}
  \label{table:expert}
\end{table}

This section report archived performance of the final network in terms of the running meters metric and manual evaluation from the sewer inspection experts.
Furthermore, analysis of the result and comment on characteristics of different types of defect are also given.
While comparison with results from relevance existing methods \cite{cheng2018automated,kumar2020deep,wang2021towards} would be useful and informative, this could not be done cleanly due to a the lack of a common train/test dataset, as mentioned in subsection \ref{subsec:sewer}.
To deal with this potential deficit, it is important to stress, that this paper leans into the current most reliable and trusted evaluators: domain experts.
On the whole, the evaluation comprises 10 sewer pipes with a total number of 1549 defects/structures detected by the experts.
According to the running meters metric, for a total of 1147 pipe sections, the network is able to produce 391 TP (34\%), 447 TN (39\%), 126 FP (11\%), and 188 FN (16\%).
In total, the accuracy is at 73.06\%.

A numerical summary of the expert's final evaluation can be seen in table \ref{table:expert}.
According to the experts, the total count of the network's detections is considerably higher than the manual detections by the experts:
frequently, network produces multiple overlapping detections for a single defect, or in other cases a single defect is covered by several non-overlapping detections.
As a rule, network's detections with very low confidence score ($<= 10\%$) are negligible false positives, and therefore are not part of the expert's evaluation.
Out of the 1549 defects/structures from the experts, 261 ($17\%$) were not found by the network and therefore are classified as false negatives. 
In table \ref{table:expert_sever}, the severity of these false negatives is classified according to \cite{dwa2007conditions}.
Most of the false negatives are classified as condition class 2 to 4, medium to minor defects with no immediate or short-term need of sanitation/renovation action; 
counting only severe false negatives the number drops to 12 out of 1549 ($0.77\%$).
This concludes the summarized evaluation result from the experts.

\begin{table}
  \centering
  \caption{Severeness of the network's false negatives. Table provided by the Dr.-Ing. Pecher und Partner Ingenieurgesellschaft mbH}
  \begin{tabular}{|r|c|l|}
    \hline
    Object count & Condition class & Severity         \\ \hline
    1            & 0               & very severe      \\ \hline
    11           & 1               & severe           \\ \hline
    159          & 2 \& 3          & med. \& slight   \\ \hline
    90           & 4               & minor            \\ \hline
  \end{tabular}
  \label{table:expert_sever}
\end{table}

This paragraph goes into analyzing the result on each type of defects/structure separately.
Possible performance affecting factors, and respective potential remedies are also presented.
First, defect types can be sorted into different categories.
Except for connection, which is a structural part of the pipe, the nine defects class can be divided into 2 subgroups, which partially explains the difference in network's reaction to each class.
These subgroups are flat (2D) defects, and spatial (3D) defects.
The difference between the two groups is, that spatial defects' one prominent feature is the offset between their surface and the normal pipe surface; 
for example, how high a pile of settled deposits is in the pipe, or how much material is gone from the inner surface of the pipe.
However, this third dimension is lost during the unrolling processing.
The flat defects group includes fissure, root.
On the other hand, spatial defects includes: angular displaced joint, settled deposit, surface damage, break/collapse, deformation, horizontal displaced joint, and obstacle
In the flat defects group, the network is able to handle root quite well.
The performance on fissure, however, still has room for improvement, specifically on finer and smaller fissures.
Within spatial defects, the network is able to detect surface damages, and obstacle quite well, despite the loss of spatial information.
Surface damages are generally easy for the network to detect due to differences of coloration and texture in comparison to healthy pipe surface.
Obstacles are also detected quite well, although it is a diverse class with multiple subclasses of very different visual characteristics, for example, encrustation, root balls, protruding shards and crossing pipes.
The first reason is that, according to the experts, they are easier to be detected in 2D images, thus less affected by the loss of spatial information.
Another possible explanation is that obstacles usually appear with other defects, .i.e root and root balls, thus detection of obstacles could be generated based on the existence of other relevant well detected classes. 
The amount of available training samples for both surface damage, and obstacles could also have positively contributed to the result. 
On the other hand, most of the spatial defects are detected poorly by the network.
The most likely cause of this is the mentioned loss of depth information.
Furthermore, these defects are also much rarer, thus having fewer training samples.
However, according to the experts, deformations and horizontal displaced joints should still be detectable without spatial data.
Thus, the more likely reason for bad performance is the lack of training data.
The settled deposit class is especially bad, despite having a somewhat sufficient amount of training samples, the network failed to find any of those at all.
Break/collapses are also in a bad position; the network often mistakes chipped connection branches or chipped joints in stone pipe for this type of defects. 
Outside of defects, the single structural class, connection, is handled easily by the network.
Due to their standardized features, connections are consistence, unambiguous, and easy to label.
In all relevant classes, connection is the closest to a typical object class in COCO or KITTI.
As a side note, the final net's $mAP@0.5$, $mAP@0.75$, and $mAP@[.5:.95]$ are 12.6, 5.5, and 5.8 respectively.

\section{\uppercase{Future Work}}
This section presents some possible ways forward, including general neural network improvements and suggested heuristics from the sewer maintenance experts.
These potential improvements can be coarsely applied to one of the following areas: the neural network and the post-processing. 
For the network itself, accuracy could be raised by adding more high quality annotated data, especially for rarer defects.
On top of that, several defect classes could be divided into more specific subclass for better differentiation; for example roots can be broken down to tap roots, independent fine roots, or complex mass of roots.
While generally sensible, this method necessitates the need for a lot more training data, and rework of current annotated dataset.
The third boost of accuracy could come from preventing the loss of spatial information during the unrolling; 
one way of achieving this would be using an RGBD camera or a stereo camera instead of a normal RGB camera.
This would greatly improve the performance on spatial defects.
Furthermore, according the experts, adding the 3rd dimension would also significantly improve performance of flat defects, such as fissures.
Another chance of improvement can also be from trying out newer SoTA network like YOLOv5, or YOLOv7 \cite{Wang2022Yolo7}, etc.; 
nonetheless, given the same dataset, there would be no guarantee of a large jump in detection quality.

During evaluation, the experts also noticed several undesirable behaviors from the network, which when rectified would greatly improve the performance and ease-of-use.
In any case, these are concrete rules and heuristics that could not be easily integrated directly into the training, because the NN training process focuses on making networks learn abstract rules from training data implicitly.
As a result, the easiest way for these explicit heuristics to be implemented would be as post-processing steps after the detection.
According to the experts, fissures and surface damages are often detected in small parts and/or multiple times.
This situation could possibly be corrected using classical methods such as Hungarian algorithm or connected components to automatically merge relevant detections together.
Another suggestion is finding a way to optimally set the minimum confidence threshold, which stems from the need to avoid overloading the experts with too many false positives.
Since each type defect are handled differently, it would also be useful to figure out one threshold for each defect type.
An additional problem with the network is, that defects on the ceiling of pipes are divided into two parts.
This is caused by trying to project a continuous cylinder into 2D space; 
while circular image convolution would be an interesting research topic, an easier way of fixing this could be using Hungarian algorithm, or similar matching algorithms to match detections on pipe's ceiling.
Finally, the following list contains several practical heuristic rules directly from the experts, which could be implemented to further better the final detection: 
\begin{itemize}
  \item	Roots can only be detected in the immediate joint and branch connection area.
  \item	Circumferential fissures are not observed in the immediate vicinity of joints, as joints pick up forces that lead to circumferential fissures at other parts of the sewer pipe.
  \item	In the joint area of vitrified clay pipe, fissures could form due to shrinkage of the glaze while the pipe is cooling down after the firing process. These fissures are only on the glaze, thus are unproblematic to pipes' structural stability.
  \item	There is a strong correlation between material of pipes, as well as locations within pipes and types of defect that would appear. For example, concrete pipes are susceptible to chemical corrosion, thus showing more surface damages over time. On the other hand, while resistant to chemical corrosion, vitrified clay pipes are brittle. Consequently, most defects found in these pipes are mechanical wear and tear, such as fissures. 
\end{itemize}
All in all, however complex and precise, post-processing still has to rely on a good detection baseline, as an extension the network.

\section{\uppercase{Conclusions}}
\label{sec:conclusion}
In this work, an EfficientDet-D0 was used to detect defects on sewer pipes' inner surface.
This network was trained on a new dataset of 14.7 km of sewer pipe, which was manually annotated by expert in the field.
At the end, the network is able to produce good detections of fissure, root, surface damage, obstacle, and connection.
However, other relevant defects with spatial feature are still difficult, due to the lack of depth information.
This problem could potentially be solved using RGB-D camera.
Furthermore, multiple post-processing using known practical heuristics could also be applied to further improve detection quality.
Finally, this work also provided some practical designs for processing and evaluating "objects" with peculiar nature, such as defects.

\section*{\uppercase{Acknowledgements}}
This work is funded by the German Federal Ministry of Education and Research under grant number 13N13891.

\bibliographystyle{apalike}
{\small
\bibliography{auzuka_arxiv}}

\end{document}